\documentclass{article}
\usepackage{ijcai20}

\usepackage{times}

\usepackage{soul}
\usepackage{url}

\usepackage[utf8]{inputenc}
\usepackage[small]{caption}
\usepackage{graphicx}
\usepackage{amsmath}
\usepackage{booktabs}
\usepackage{microtype}
\usepackage{amsthm}
\usepackage{amsmath}
\usepackage{inconsolata}
\usepackage{listings}
\usepackage{xcolor}
\usepackage[bitstream-charter]{mathdesign}

\newcommand{\metaho}{Metagol$_{ho}$}
\newcommand{\metabias}{Metagol$_{DF}$}
\newcommand{\forgetgol}{Forgetgol}
\newcommand{\dilp}{$\partial$ILP}
\newcommand{\tw}[1]{\texttt{#1}}

\theoremstyle{definition}
\newtheorem{example}{Example}

\title{Turning 30: New Ideas in Inductive Logic Programming}

\author{
Andrew Cropper$^1$\and
Sebastijan Dumančić$^2$,\And
Stephen H. Muggleton$^3$\\
\affiliations
$^1$University of Oxford\\
$^2$KU Leuven\\
$^3$Imperial College London\\
\emails
andrew.cropper@cs.ox.ac.uk,
sebastijan.dumancic@cs.kuleuven.be,
stephen.muggleton@imperial.ac.uk
}

\begin{document}

\maketitle

\begin{abstract}
Common criticisms of state-of-the-art machine learning include poor generalisation, a lack of interpretability, and a need for large amounts of training data.
We survey recent work in inductive logic programming (ILP), a form of machine learning that induces logic programs from data, which has shown promise at addressing these limitations.
We focus on new methods for learning recursive programs that generalise from few examples, a shift from using hand-crafted background knowledge to \emph{learning} background knowledge, and the use of different technologies, notably answer set programming and neural networks.
As ILP approaches 30, we also discuss directions for future research.
\end{abstract}\section{Introduction}
Common criticisms of state-of-the-art machine learning include poor generalisation, a lack of interpretability, and a need for large numbers of training examples \cite{marcus:2018,chollet:2019,bengio:2019}.
In this paper, we survey recent work in inductive logic programming (ILP) \cite{mugg:ilp}, a form of machine learning that induces logic programs from data, which has shown promise at addressing these limitations.

Compared to most machine learning approaches, ILP has several advantages.
ILP systems can learn using background knowledge (BK) (relational theories), such as using a theory of light to understand images \cite{mugg:vision}.
Because of the expressivity of logic programs, ILP systems can learn complex relational theories, such as cellular automata \cite{inoue:lfit}, event calculus theories \cite{oled}, and Petri nets \cite{DBLP:journals/ml/BainS18}.
Because hypotheses are logic programs, they can be read by humans, crucial for explainable AI and ultra-strong machine learning \cite{michie:usml}.
Due to the strong inductive bias imposed by the BK, ILP systems can generalise from small numbers of examples, often a single example \cite{mugg:metabias}.
Finally, because of their symbolic nature, ILP systems naturally support lifelong and transfer learning \cite{crop:forget}, which is considered essential for human-like AI \cite{lake:ai}.


\begin{table}[ht]
\centering
\footnotesize
\begin{tabular}{@{}lll@{}}
\toprule
& \textbf{Old ILP} & \textbf{New ILP} \\
\midrule
\textbf{Recursion} & Limited & Yes\\
\textbf{Predicate invention} & No & Yes\\
\textbf{Hypotheses} & First-order & Higher-order, ASP\\
\textbf{Optimality} & No & Yes\\
\textbf{Technology} & Prolog & Prolog, ASP, NNs \\
\bottomrule
\end{tabular}
\caption{
    A simplified comparison between old and new ILP systems.
}
\label{tab:diffs}
\end{table}

Some of these advantages come from recent developments, which we survey in this paper.
To aid the reader, we coarsely compare old and new ILP systems.
We use FOIL \cite{foil}, Progol \cite{progol}, TILDE \cite{tilde}, and Aleph \cite{aleph} as representative old systems and ILASP \cite{law:ilasp}, Metagol \cite{metagol}, and \dilp{} \cite{evans:dilp} as representative new systems.
This comparison, shown in Table \ref{tab:diffs}, is, of course, vastly over simplified, and there are many exceptions to the comparison.
We discuss each development (each row in the table) in turn.
We discuss new methods for learning recursive logic programs, which allows for more generalisation from fewer examples (Section \ref{sec:programs});
a shift from using hand-crafted BK to learning BK, namely through \emph{predicate invention} and \emph{transfer learning}, which has been shown to improve predictive accuracies and reduce learning times (Section \ref{sec:background});
learning programs of various expressivity, notably Datalog and answer set programs (Section \ref{sec:expr});
new methods for learning optimal programs, such as efficient time complexity programs (Section \ref{sec:optimal});
and recent work that uses different underlying technologies to perform the learning, notably to leverage recent improvements with ASP/SMT solvers and neural networks (Section \ref{sec:tech}).
Finally, as ILP approaches 30, we conclude by proposing directions for future research.

\subsection{What is ILP?}

Given positive and negative examples and BK, the goal of an ILP system is to induce (learn) a hypothesis (a logic program) which, with the BK, entails as many positive and as few negative examples as possible \cite{mugg:ilp}.
Whereas most machine learning approaches represent data as vectors, ILP represents data (examples, BK, and hypotheses) as logic programs.

\begin{example}
\label{ex:intro}

To illustrate ILP, suppose you want to learn a string transformation program from the following examples.

\begin{center}
\begin{tabular}{@{}c|c@{}}
\toprule
\textbf{Input} & \textbf{Output} \\
\midrule
machine & e \\
learning & g \\
algorithm & m \\
\bottomrule
\end{tabular}
\end{center}

Most forms of machine learning would represent these examples as vectors.
By contrast, in ILP, we would represent these examples as logical atoms, such as \tw{f([m,a,c,h,i,n,e], e)}, where \tw{f} is the target predicate that we want to learn (i.e. the relation to generalise).
BK (features) is likewise represented as a logical theory (a logic program).
For instance, for the string transformation problem, we could provide BK that contains logical definitions for string operations, such as
\tw{empty(A)}, which holds when the list \tw{A} is empty,
\tw{head(A,B)}, which holds when \emph{B} is the head of the list \tw{A},
and \tw{tail(A,B)}, which holds when \tw{B} is the tail of the list \tw{A}.
Given the aforementioned examples and BK, an ILP system could induce the hypothesis (a logic program) shown in Figure \ref{fig:intro-prog}.

\begin{figure}[ht]
\begin{center}
\begin{tabular}{|c|}
\hline
\begin{lstlisting}
f(A,B):-tail(A,C),empty(C),head(A,B).
f(A,B):-tail(A,C),f(C,B).
\end{lstlisting} \\
\hline
\end{tabular}
\end{center}
\caption{
    A hypothesis (logic program) for the string transformation problem in  Example \ref{ex:intro}.
    Each line of the program is a logical clause (or rule).
    The first rule says that the relation \tw{f(A,B)} holds when the three literals \tw{tail(A,C)}, \tw{empty(C)}, and \tw{head(A,B)} hold, i.e. \tw{B} is the last element of \tw{A} when the tail of \tw{A} is empty and \tw{B} is the head of \tw{A}.
    The second rule is recursive and says that the relation \tw{f(A,B)} holds when the two literals \tw{tail(A,C)} and \tw{f(C,B)} hold, i.e. \tw{f(A,B)} holds when the same relation holds for the tail of \tw{A}.
}
\label{fig:intro-prog}
\end{figure}
\end{example}

\section{Recursion}
\label{sec:programs}


A recursive logic program is one where the same predicate appears in the head and body of a rule.
Learning recursive programs has long been considered a difficult problem for ILP \cite{mlj:ilp20}.
To illustrate the importance of recursion, consider the string transformation problem in Example \ref{ex:intro}.
With recursion, an ILP system can learn the compact program shown in Figure \ref{fig:intro-prog}.
Because of the symbolic representation and the recursive nature, the program generalises to lists of arbitrary size and to lists that contain arbitrary elements (e.g integers, characters, etc).
Without recursion, an ILP system would need to learn a separate clause to find the last element for each list of length $n$, such as:

\begin{center}
\begin{tabular}{|c|}
\hline
\begin{lstlisting}[basicstyle=\ttfamily\scriptsize,captionpos=b,columns=flexible]
f(A,B):-tail(A,C),empty(C),head(A,B).
f(A,B):-tail(A,C),tail(C,D),empty(D),head(C,B).
f(A,B):-tail(A,C),tail(C,D),tail(D,E),empty(E),head(E,B).
\end{lstlisting} \\
\hline
\end{tabular}
\end{center}

In other words, without recursion, it is often difficult for an ILP system to generalise from small numbers of examples \cite{crop:datacurate}.



Older ILP systems struggle to learn recursive programs, especially from small numbers of training examples.
A common limitation with existing approaches is that they rely on \emph{bottom clause construction} \cite{progol}.
In this approach, for each example, an ILP system creates the most specific clause that entails the example, and then tries to generalise the clause to entail other examples.
However, this sequential covering approach requires examples of both the base and inductive cases, and in that order.

Interest in recursion resurged with the introduction of meta-interpretive learning (MIL) \cite{mugg:metalearn,mugg:metagold,crop:metaho} and the MIL system Metagol \cite{metagol}.
The key idea of MIL is to use metarules, or program templates, to restrict the form of inducible programs, and thus the hypothesis space\footnote{
    The idea of using metarules to restrict the hypothesis space has been widely adopted by many approaches \cite{wang2014structure,AlbarghouthiDatalog2017,DBLP:conf/nips/Rocktaschel017,evans:dilp,SiDatalog2018,DBLP:journals/ml/BainS18,DBLP:conf/ijcai/SiRHN19,hexmil}.
    However, despite their now widespread use, there is little work determining which metarules to use for a given learning task (\cite{crop:reduce} is an exception), which future work must address.
}.
A metarule is a higher-order clause.
For instance, the \emph{chain} metarule is $P(A,B) \leftarrow Q(A,C), R(C,B)$, where the letters $P$, $Q$, and $R$ denote higher-order variables and $A$, $B$ and $C$ denote first-order variables.
The goal of a MIL system, such as Metagol, is to find substitutions for the higher-order variables.
For instance, the \emph{chain} metarule allows Metagol to induce programs such as \tw{f(A,B):-tail(A,C),head(C,B)}\footnote{Metagol can induce longer clauses though predicate invention, which is described in Section \ref{sec:pi}.}.
Metagol induces recursive programs using recursive metarules, such as the \emph{tail recursion} metarule \emph{P(A,B) $\leftarrow$ Q(A,C), P(C,B)}.

Following MIL, many ILP systems can learn recursive programs \cite{law:ilasp,evans:dilp,hexmil}.
With recursion, ILP systems can now generalise from small numbers of examples, often a single example \cite{mugg:metabias,crop:playgol}.
The ability to learn recursive programs has opened up ILP to new application areas, including learning string transformations programs \cite{mugg:metabias}, robot strategies \cite{crop:robots}, and answer set grammars \cite{law:asg}.

\section{Learning Background Knowledge}
\label{sec:background}

A key characteristic of ILP is the use of BK as a form of inductive bias.
BK is similar to features used in most forms of machine learning.
However, whereas features are vectors, BK usually contains facts and rules (extensional and intensional definitions) in the form of a logic program.
For instance, when learning string transformation programs, we may want to supply helper background relations, such as \tw{head/2}\footnote{Notation for a predicate symbol \tw{head} with two arguments.} and \tw{tail/2}.
For other domains, we may want to supply more complex BK, such as a theory of light to understand images \cite{mugg:vision} or higher-order operations, such as \tw{map/3}, \tw{filter/3}, and \tw{fold/4}, to solve programming puzzles \cite{crop:metaho}.

As with choosing appropriate features, choosing appropriate BK is crucial for good learning performance.
ILP has traditionally relied on hand-crafted BK, often designed by domain experts, i.e. feature engineering.
This approach is clearly limited because obtaining suitable BK can be difficult and expensive.
Indeed, the over reliance on hand-crafted BK is a common criticism of ILP \cite{evans:dilp}.

Two recent avenues of research attempt to overcome this limitation.
The first idea is to enable an ILP system to automatically \emph{invent} new predicate symbols.
The second idea is to perform lifelong and transfer learning to discover knowledge that can be reused to help learn other programs.
We discuss these ideas in turn.


\subsection{Predicate Invention}
\label{sec:pi}

Rather than expecting a user to provide all the necessary BK, the goal of \emph{predicate invention} is for an ILP system to automatically invent new auxiliary predicate symbols.
This idea is similar to when humans create new functions when manually writing programs, as to reduce code duplication or to improve readability.

Whilst predicate invention has attracted interest since the beginnings of ILP \cite{mugg:cigol}, most previous attempts have been unsuccessful resulting in no support for predicate invention in popular ILP systems \cite{foil,progol,tilde,aleph}.
A key limitation of early ILP systems is that the search is complex and under-specified in various ways.
For instance, it was unclear how many arguments an invented predicate should have, and how the arguments should be ordered.

As with recursion, interest in predicate invention has resurged with the introduction of MIL.
MIL avoids the issues of older ILP systems by using metarules to define the hypothesis space and in turn reduce the complexity of inventing a new predicate symbol.
As mentioned, a metarule is a higher-order clause.
For instance, the \emph{chain} metarule ($P(A,B) \leftarrow Q(A,C), R(C,B)$) allows Metagol to induce programs such as \tw{f(A,B):- tail(A,C),tail(C,D)}, which would drop the first two elements from a list.
To induce longer clauses, such as to drop first three elements from a list, Metagol can use the same metarule but can invent a new predicate symbol and then chain their application, such as to induce the program:

\begin{center}
\begin{tabular}{|c|}
\hline
\begin{lstlisting}
f(A,B):-tail(A,C),inv1(C,B).
inv1(A,B):-tail(A,C),tail(C,B).
\end{lstlisting} \\
\hline
\end{tabular}
\end{center}

To learn this program, Metagol invents the predicate symbol \tw{inv1} and induces a definition for it using the \emph{chain} metarule.
Metagol uses this new predicate symbol in the definition for the target predicate \tw{f}.

Predicate invention allows Metagol (and other ILP systems) to learn programs by expanding their BK.
A major side-effect of this metarule and predicate invention driven approach is that problems are forced to be decomposed into smaller problems, where the decomposed solutions can be reused.
For instance, suppose you wanted to learn a program that drops the first four elements of a list, then Metagol could learn the following program, where the invented predicate symbol \tw{inv1} is used twice:

\begin{center}
\begin{tabular}{|c|}
\hline
\begin{lstlisting}
f(A,B):-inv1(A,C),inv1(C,B).
inv1(A,B):-tail(A,C),tail(C,B).
\end{lstlisting} \\
\hline
\end{tabular}
\end{center}

Predicate invention has been shown to help reduce the size of target programs, which in turns reduces sample complexity and improves predictive accuracy \cite{crop:playgol,crop:metaho}.
Following Metagol, other newer ILP systems support predicate invention \cite{evans:dilp,hexmil}, often using a metarule guided approach.
Such systems all have the general principle of introducing new predicate symbols when their current BK is insufficient to learn a hypothesis.

In contrast to aforementioned approaches, a different idea is to invent new predicates to improve knowledge representation.
For instance, CUR$^2$LED \cite{curled} learns new predicates by clustering constants and relations in the provided BK, turning each identified cluster into a new BK predicate.
The key insight of CUR$^2$LED is not to use a single similarity measure, but rather a set of various similarities.
This choice is motivated by the fact that different similarities are useful for different tasks, but in the unsupervised setting the task itself is not known in advance.
CUR$^2$LED would therefore invent predicates by producing different clusterings according to the features of the objects, community structure and so on.

ALPs \cite{dumancic:encoding} perform predicate invention inspired by a (neural) auto-encoding principle: they learn an \emph{encoding} logic program that maps the provided data to a new, compressive latent representation (defined in terms of the invented predicates), and a \emph{decoding} logic program that can reconstruct the provided data from its latent representation.
Both approaches have demonstrated an improved performance on supervised tasks, even though the predicate invention step is task-agnostic.

\subsection{Lifelong Learning}
\label{sec:lifelong}

An advantage of a symbolic representation is that learned knowledge can be remembered, i.e. explicitly stored in the BK.
Therefore, the second line of research that tries to address the limitations of hand-crafted BK tries to leverage transfer learning.
The general idea is to reuse knowledge gained from solving one problem to help solve a different problem.


One notable application of transfer learning is the \metabias{} system \cite{mugg:metabias} which, given a set of tasks, uses Metagol to try to learn a solution for each task using at most 1 clause.
If Metagol finds a solution for any task, it adds the solution to the BK and removes the task from set.
It then tries to find solutions for the rest of the tasks, but can now (1) use an additional clause, and (2) reuse solutions from solved tasks.
This process repeats until \metabias{} solves all the tasks, or reaches a maximum program size.
In other words, \metabias{} automatically identifies easier problems, learn programs for them, and then reuses the solutions to help learn programs for more difficult problems.
One of the key ideas of Metabias is to not only save the induced target relation to the BK, but to also add its constituent parts discovered through predicate invention.
The authors experimentally show that their multi-task approach performs substantially better than a single-task approach because learned programs are frequently reused.
Moreover, they show that this approach leads to a hierarchy of BK composed of reusable programs, where each builds on simpler programs, which can be seen as \emph{deep inductive logic programming}.

\metabias{} saves all learned programs (including invented predicates) to the BK, which can be problematic because too much irrelevant BK is detrimental to learning performance.
To address this problem, \forgetgol{} \cite{crop:forget} introduces the idea of \emph{forgetting}.
In this approach, \forgetgol{} continually grows and shrinks its hypothesis space by adding and removing learned programs to and from its BK.
The authors show that forgetting can reduce both the size of the hypothesis space and the sample complexity of an ILP learner when learning from many tasks, which shows potential for ILP to be useful in a lifelong or continual learning setting, which is considered crucial for AI \cite{lake:ai}.



The aforementioned \metabias{} and \forgetgol{} approaches assume a corpus of user-supplied tasks to train from.
This assumption is unrealistic in many situations.
To overcome this limitation, Playgol \cite{crop:playgol} first \emph{plays} by randomly sampling its own tasks to solve, and tries to solve them, adding any solutions to the BK, which can be seen as a form of \emph{self-supervised} learning.
After playing Playgol tries to solve the user-supplied tasks by reusing solutions learned whilst playing.
The goal of Playgol is similar to all the approaches discussed in this section: to automatically discover reusable general programs as to improve learning performance, but does so with fewer labelled examples.



\section{Expressiveness}
\label{sec:expr}

ILP systems have traditionally induced Prolog programs.
A recent development has been to use alternative hypothesis representations.

\subsection{Datalog}
Datalog is a syntactical subset of Prolog which disallows complex terms as arguments of predicates and imposes restrictions on the use of negation (and negation with recursion).
These restrictions make Datalog attractive for two reasons.
First, Datalog is a truly declarative language, whereas in Prolog reordering clauses can change the program.
Second, a Datalog query is guaranteed to terminate, though this guarantee is at the expense of not being a Turing-complete language, which Prolog is.
Due to the aforementioned benefits, several works \cite{AlbarghouthiDatalog2017,evans:dilp,SiDatalog2018,RaghothamanDatalog2019} induce Datalog programs.
The general motivation for reducing the expressivity of the representation language from Prolog to Datalog is to allow the problem to be encoded as a satisfiability problem, particularly to leverage recent developments in SAT and SMT.
We discuss the advantages of this approach more in Section \ref{sec:sat}.


\subsection{Answer Set Programming}
\label{sec:hyp:asp}
ASP is a logic programming paradigm.
Like Datalog, ASP is a truly declarative language.
Compared to Datalog, ASP is more expressive, allowing, for instance, disjunction in the head of a clause, hard and weak constraints, and support for default inference and default assumptions.
A key difference between ASP and Prolog is semantics.
A definite logic program (a Prolog program) has only one model (the least Herbrand model).
By contrast, an ASP program can have one, many, or even no models (answer sets).
Due to its non-monotonicity, ASP are particularly attractive for expressing common-sense reasoning \cite{DBLP:journals/ai/LawRB18}.



Approaches to learning ASP programs can mostly be divided into two categories: \emph{brave learners}, which aim to learn a program such that at least one answer set covers the examples, and \emph{cautious learners}, which aim to find a program which covers the examples in all answer sets.
ILASP (Inductive Learning of Answer Set Programs) \cite{law:ilasp} is a collection of ILP systems that learn ASP programs.
ILASP is notable because it supports both brave and cautious learning, which are both needed to learn some ASP programs \cite{DBLP:journals/ai/LawRB18}.
Moreover, ILASP differs from most Prolog-based ILP systems because it learns unstratified ASP programs, including programs with normal rules, choice rules, and both hard and weak constraints, which classical ILP systems cannot.
Learning ASP programs allows for ILP to be used for new problems, such as inducing answer set grammars \cite{law:asg}.


\subsection{Higher-Order Programs}

Imagine learning a \emph{droplasts} program, which removes the last element of each sublist in a list, e.g. \emph{[alice,bob,carol]} $\mapsto$ \emph{[alic,bo,caro]}.
Given suitable input data, Metagol can learn this first-order recursive program:

\begin{center}
\begin{tabular}{|c|}
\hline
\begin{lstlisting}
f(A,B):-empty(A),empty(B).
f(A,B):-head(A,C),tail(A,D),head(B,E),
        tail(B,F),f1(C,E),f(D,F).
f1(A,B):-reverse(A,C),tail(C,D),reverse(D,B).
\end{lstlisting} \\
\hline
\end{tabular}
\end{center}

Although semantically correct, the program is verbose.
To learn more compact programs, \metaho{} \cite{crop:metaho} extends Metagol to support learning higher-order programs, where predicate symbols can be used as terms.
For instance, for the same \emph{droplasts} problem, \metaho{} learns the higher-order program:

\begin{center}
\begin{tabular}{|c|}
\hline
\begin{lstlisting}
f(A,B):-map(A,B,f1).
f1(A,B):-reverse(A,C),tail(C,D),reverse(D,B).
\end{lstlisting} \\
\hline
\end{tabular}
\end{center}

To learn this program, \metaho{} invents the predicate symbol \tw{f1}, which is used twice in the program: once as term in the \tw{map(A,B,f1)} literal and once as a predicate symbol in the \tw{f1(A,B)} literal.
Compared to the first-order program, this higher-order program is smaller because it uses \tw{map/3} to abstract away the manipulation of the list and to avoid the need to learn an explicitly recursive program (recursion is implicit in \tw{map/3}).
By reducing the size of the target program by learning higher-order programs, \metaho{} has been shown to reduce sample complexity and learning times, and improve predictive accuracies \cite{crop:metaho}.
This example illustrates the value of higher-order abstractions and inventions, which allow ILP systems to learn more complex programs using fewer examples with less search.


%
%

\section{Optimal Programs}
\label{sec:optimal}

In ILP there are often multiple (sometimes infinite) hypotheses that explain the data.
Deciding which hypothesis to choose has long been a difficult problem.
Older ILP systems were not guaranteed to induce optimal programs, where optimal typically means with respect to the size of the induced program, or the coverage of examples.

A key reason for this limitation was that most search techniques learned a single clause at a time, leading to the construction of sub-programs which were sub-optimal in terms of program size and coverage.
For instance, programs induced by Aleph offer no guarantee of optimality with respect to the program size and coverage.

Newer ILP systems try to address this limitation.
As with the ability to learn recursive programs, the main development is to take a global view of the induction task.
In other words, rather than induce a single clause at a time from a subset of the examples, the idea is to induce a whole program.
For instance, ILASP is given as input a hypothesis space with a set of candidate clauses.
The ILASP task is to find a minimal subset of clauses that covers as many positive and as few negative examples as possible.
To do so, ILASP uses ASP's optimisation abilities to provably learn the program with the fewest literals.
Likewise, Metagol and HEXMIL are guaranteed to induce programs with the fewest clauses.

An advantage of learning optimal programs is learning performance.
The idea is that the smallest program should provide better generalisations.
When \citeauthor{law:noisy} (\citeyear{law:noisy}) compared ILASP (which is guaranteed to learn optimal programs) to Inspire \cite{inspirecomp} (which is not guaranteed to learn optimal programs), ILASP achieved a higher F1 score (both systems were given identical hypothesis spaces and optimisation criteria).

In addition to performance advantages, the ability to learn optimal programs opens up ILP to new problems.
For instance, learning efficient logic programs has long been considered a difficult problem in ILP \cite{mugg:ilp94,mlj:ilp20}, mainly because there is no declarative difference between an efficient program, such as mergesort, and an inefficient program, such as bubble sort.
To address this issue, Metaopt \cite{crop:metaopt} extends Metagol to support learning efficient programs.
Metaopt maintains a cost during the hypothesis search and uses this cost to prune the hypothesis space.
To learn minimal time complexity logic programs, Metaopt minimises the number of resolution steps.
For instance, imagine trying to learn a \emph{find duplicate} program, which finds any duplicate element in a list e.g. \emph{[p,r,o,g,r,a,m]} $\mapsto$ \emph{r}, and \emph{[i,n,d,u,c,t,i,o,n]
} $\mapsto$ \emph{i}.
Given suitable input data, Metagol can induce the recursive program:

\begin{center}
\begin{tabular}{|c|}
\hline
\begin{lstlisting}
f(A,B):-head(A,B),tail(A,C),element(C,B).
f(A,B):-tail(A,C),f(C,B).
\end{lstlisting} \\
\hline
\end{tabular}
\end{center}

This program goes through the elements of the list checking whether the same element exists in the rest of the list.
Given the same input, Metaopt induces the recursive program:

\begin{center}
\begin{tabular}{|c|}
\hline
\begin{lstlisting}
f(A,B):-mergesort(A,C),f1(C,B).
f1(A,B):-head(A,B),tail(A,C),head(C,B).
f1(A,B):-tail(A,C),f1(C,B).
\end{lstlisting} \\
\hline
\end{tabular}
\end{center}

This program first sorts the input list and then goes though the list to check whether for duplicate adjacent elements.
Although larger, both in terms of clauses and literals, the program learned by Metaopt is more efficient $O(\log n)$ than the program learned by Metagol $O(n^2)$.
The main implication of this work is that Metaopt can learn efficient robot strategies, efficient time complexity logic programs, and even efficient string transformation programs.

Following this idea, FastLAS \cite{law:fastlas} is an ASP-based ILP system that takes as input a custom scoring function and computes an optimal solution with respect to the given scoring function.
The authors show that this approach allows a user to optimise domain-specific performance metrics on real-world datasets, such as access control policies.\section{Different Technologies}
\label{sec:tech}

Older ILP systems mostly use Prolog for reasoning.
Recent work considers using different technologies.

\subsection{Constraint Satisfaction and Satisfiability}
\label{sec:sat}
There have been tremendous recent advances in SAT and SMT solvers.
To leverage these advances, much recent work uses ASP to induce logic programs \cite{xhail,aspal,mugg:metalearn,law:ilasp,iled,oled,hexmil}.
The main motivations are to leverage (1) the language benefits of ASP (Section \ref{sec:hyp:asp}), and (2) the efficiency and optimisation techniques of modern ASP solvers, such as CLASP \cite{clasp}, which supports conflict propagation and learning.
With similar motivations, several works \cite{AlbarghouthiDatalog2017,SiDatalog2018,RaghothamanDatalog2019} synthesise Datalog program by encoding the ILP task into a SMT problem.

These approaches have been shown able to reduce learning times compared to standard Prolog-based approaches.
However, some unresolved issues remain.
A key issue is that most approaches encode an ILP problem as a single (often very large) satisfiability problem.
These approaches therefore often struggle to scale to very large problems \cite{crop:metaho}.
Although preliminary work attempts to tackle this issue \cite{law:fastlas}, work is still needed for these approaches to scale to large problems.


%


\subsection{Neural Networks}
With the recent rise of deep learning and neural networks, several approaches have explored using gradient-based methods to learn logic programs.
These approaches all replace absolute logical reasoning with a relaxed version that yields continuous values reflecting the confidence of the conclusion.
Although this approach limits the expressivity of hypotheses, it potentially allows for gradient-based methods to be used to learn from large datasets.

The research has primarily developed in three directions.
The first concerns imitating logical reasoning with tensor calculus \cite{Cohen_NeuralLP,NLM}.
These approaches represent predicates as binary tensors over the domain of constants and perform reasoning by chains of tensor products imitating as clause.
The second concerns the relaxation of the subset selection problem \cite{evans:dilp,DBLP:conf/ijcai/SiRHN19} in which the task of a neural network is to select a subset of clauses from a space of pre-defined clauses.
The third, \emph{neural theorem provers} \cite{DBLP:conf/nips/Rocktaschel017} turn the learning problem towards learning to perform \emph{soft} unification, which unifies not only the matching symbols but also similar ones, from a fixed set of proofs.

The major challenge of neural approaches is the inability to generalise beyond training data and data efficiency.
The majority of these approaches \textit{embed} logical symbols, i.e. they replace symbols with vectors, and therefore  a learned model is unable to work with unseen constants.
Moreover, neural methods often require millions of examples \cite{NLM} to learn concepts that symbolic ILP is able to learn from just a few.


%
%
%



\section{Limitations and Future Work}
The recent advances surveyed in this paper have opened new problems for future work to address.

\paragraph{Relevance}
New methods for predicate invention (Section \ref{sec:pi}) and transfer learning (Section \ref{sec:lifelong}) have improved the abilities of ILP systems to learn large programs.
Moreover, these techniques raise the potential for ILP to be used in lifelong learning settings.
However, inventing and acquiring new BK could lead to a problem of too much BK, which can overwhelm an ILP system \cite{crop:forget}.
On this issue, a key under-explored topic is that of \emph{relevancy}.
Given a new induction problem with large amounts of BK, how does an ILP system decide which BK is relevant?
Without efficient methods of relevance identification, it is unclear how efficient lifelong learning can be achieved.


\paragraph{Noisy BK}
\label{noisy}
Lifelong learning is seen as key to AI, and recent work in ILP has shown promise in this direction (Section \ref{sec:lifelong}).
However, unresolved issues remain.
One key issue is the underlying uncertainty associated with adding learned programs to the BK.
By the nature of induction, such programs are expected to be noisy, yet they are the building blocks for further inductive inference.
Building noisy programs on top of other noisy programs could lead to eventual incoherence of the learned program.

\paragraph{Probabilistic ILP}
A principled way to handle noise is to unify logical and probabilistic reasoning, which is the focus of \emph{statistical relational artificial intelligence} (StarAI) \cite{DeRaedtKerstingEtAl16}.
While StarAI is a growing field, inducing probabilistic logic programs has received little attention, with few notable exceptions \cite{DBLP:journals/tplp/BellodiR15,probfoil}, as inference remains the main challenge.
Addressing this issue, i.e. unifying probabiliy and logic in an inductive setting, would be a major achievement \cite{marcus:2018} and the ILP developments outlined in this paper will be a crucial element of the progress.

\paragraph{Explainability}
\label{explain}
Explainability is one of the claimed advantages of a symbolic representation.
Recent work \cite{mugg:compmlj} evaluates the comprehensibility of ILP hypotheses using Michie's (\citeyear{michie:usml}) framework of \emph{ultra-strong machine learning}, where a learned hypothesis is expected to not only be accurate but to also demonstrably improve the performance of a human being provided with the learned hypothesis.
The paper empirically demonstrates improved human understanding directly through learned hypotheses.
However, more work is required to better understand the conditions under which this can be achieved.

\paragraph{Summary}
As ILP approaches 30, we think that the recent advances surveyed in this paper puts ILP in a prime position to have a significant impact on AI over the next decade, especially to address the key limitations of state-of-the-art machine learning.

{
\small
\bibliographystyle{named}
\bibliography{ourbib15}
}
\end{document}